\documentclass{ecai} 
\usepackage{amsmath}
\usepackage{graphicx}
\usepackage{xcolor} 
\usepackage{booktabs}
\usepackage{multirow}
\usepackage{makecell} % da mettere nel preambolo

\usepackage{latexsym}
\usepackage{amssymb}
\usepackage{amsmath}
\usepackage{amsthm}
\usepackage{booktabs}
\usepackage{enumitem}
\usepackage{graphicx}
\usepackage{color}

\newcommand{\BibTeX}{B\kern-.05em{\sc i\kern-.025em b}\kern-.08em\TeX}

\begin{document}

%%%%%%%%%%%%%%%%%%%%%%%%%%%%%%%%%%%%%%%%%%%%%%%%%%%%%%%%%%%%%%%%%%%%%%%%

\begin{frontmatter}

%%% Use this command to specify your submission number.
%%% In doubleblind mode, it will be printed on the first page.

\paperid{2637} 

%%% Use this command to specify the title of your paper.

\title{Leveraging Large Language Models for Accurate Sign Language Translation in Low-Resource Scenarios}

%%% Use this combinations of commands to specify all authors of your 
%%% paper. Use \fnms{} and \snm{} to indicate everyone's first names 
%%% and surname. This will help the publisher with indexing the 
%%% proceedings. Please use a reasonable approximation in case your 
%%% name does not neatly split into "first names" and "surname".
%%% Specifying your ORCID digital identifier is optional. 
%%% Use the \thanks{} command to indicate one or more corresponding 
%%% authors and their email address(es). If so desired, you can specify
%%% author contributions using the \footnote{} command.
\author[A,B]{\fnms{Luana}~\snm{Bulla}} %\orcid{0000-0003-1165-853X}}
\author[A,B]{\fnms{Gabriele}~\snm{Tuccio}\thanks{Corresponding Author. Email: gabriele.tuccio@phd.unict.it}} %orcid{0009-0002-5224-9599}
\author[A,B]{\fnms{Misael}~\snm{Mongiovì}} %\orcid{0000-0003-0528-5490}
\author[B,C]
{\fnms{Aldo}~\snm{Gangemi}}%\footnote{Equal contribution.}} %\orcid{0000-0001-5568-2684}

%\footnotemark

\address[A]{University of Catania, Italy}
\address[B]{National Research Council - ISTC, Italy }
\address[C]{University of Bologna, Italy}

%%% Use this environment to include an abstract of your paper.

\begin{abstract}
Translating natural languages into sign languages is a highly complex and underexplored task.
Despite growing interest in accessibility and inclusivity, the development of robust translation systems remains hindered by the limited availability of parallel corpora which align natural language with sign language data. Existing methods often struggle to generalize in these data-scarce environments, as the few datasets available are typically domain-specific, lack standardization, or fail to capture the full linguistic richness of sign languages.
To address this limitation, we propose Advanced Use of LLMs for Sign Language Translation (AulSign), a novel method that leverages Large Language Models via dynamic prompting and in-context learning with sample selection and subsequent sign association. 
Despite their impressive abilities in processing text, LLMs lack intrinsic knowledge of sign languages; therefore, they are unable to natively perform this kind of translation. To overcome this limitation, we associate the signs with compact descriptions in natural language and instruct the model to use them.  
We evaluate our method on both English and Italian languages using SignBank+, a recognized benchmark in the field, as well as 
the Italian LaCAM CNR-ISTC dataset.
We 
% specific dataset based on SignPuddle, and
demonstrate superior performance compared to state-of-the-art models in low-data scenario. Our findings demonstrate the effectiveness of AulSign, with the potential to enhance accessibility and inclusivity in communication technologies for underrepresented linguistic communities.\footnote{} \end{abstract}
\end{frontmatter}

\section{Introduction}
\label{sect:intro}

Automatic translation has seen significant advancements with Large Language Models (LLMs)~\cite{hadi2024large,wang2023document,zhang2023prompting}. These models, often comprising billions of parameters, excel at translating widely spoken languages. However, their performance declines significantly for low-resource languages and domain-specific text formats~\cite{daniel2024impact}.
This limitation stems from their reliance on training data dominated by widely spoken languages, which constrains their ability to understand, represent, and translate underrepresented linguistic systems.
Among low-resource languages, sign languages present significant challenges. Sign languages, such as Italian Sign Language (LIS) and American Sign Language (ASL), rely on a visual-spatial grammar system rather than spoken or written syntax. This linguistic structure, coupled with the scarcity of available training data,
makes accurate translation more difficult to achieve.
Furthermore, translating spoken language into sign language is under-explored in the current research.
Addressing this challenge demands effective methods in data-scarce scenarios, leveraging external linguistic resources like specialized vocabularies and lexicons to produce coherent and explainable translations.

Sign Language Translation (SLT) is usually approached by separating the graphical part (computer vision or video generation) from the language part (translation) and use an intermediate language for encoding the sign language. 
% One common representation is based on glosses. Despite its simplicity, gloss-based representations tend to oversimplify the linguistic complexity of sign languages. 
% 
% Sign languages are fully developed natural languages that employ a complex system of manual gestures, facial expressions, and body movements to convey meaning. Unlike spoken languages, they are inherently visual-spatial, relying on three-dimensional space to express grammatical and semantic information. This unique modality poses significant challenges for linguistic and computational analysis.
Common intermediate representations are gloss notation, HamNoSys~\cite{prillwitz1990hamburg}, and SignWriting~\cite{sutton2022lessons}.
% , have been developed to bridge these gaps. 
Glosses provide simple written approximations using natural language labels, while HamNoSys offers a phonetic transcription of handshapes, movements, and locations in a language-independent manner. However, these systems often fail to fully capture the holistic visual-spatial structure of signed communication.
SignWriting addresses these limitations by graphically encoding key features of signs (e.g., handshapes, orientations, movements, body locations) within a two-dimensional layout that mirrors the structure of signed expressions. 
In addition to symbolic systems, recent approaches have explored keypoint-based representations such as SMPL-X~\cite{pavlakos2019expressive}, which parametrically model the body, hands, and face in a continuous space, enabling direct animation of virtual avatars and bypassing the need for manual symbolic annotation. All described approaches require the availability of large amount of training data, which are not always available, especially for under-represented sign languages or in specific domains, e.g. education.  

We propose Advanced Use of LLMs for Sign Language Translation (AulSign), a novel sign language translation method that can handle languages not well represented in the LLM training data.\footnote{This paper has been published to ECAI 2025~\cite{bulla2025language}. Code and supplemental material are available on \url{https://github.com/gabrix00/AulSign}}
Our method leverages formalized lexicons specific to a given domain, incorporating external vocabularies to enhance translation. 
By employing in-context learning via few-shot prompting,  our method significantly reduces the need for large training corpora, while introducing a novel, transparent, and explainable translation process at each stage.
Our method, which addresses both spoken-to-sign and sign-to-spoken translation tasks, comprises three core components: a Retriever, an LLM and a Sign Mapper. 
The Retriever module identifies and retrieves samples from a training set, which are used to instruct the LLM to map the input sentence into a pseudo-language that represents the sign language as a sequence of univocal descriptions of signs (we refer to a pseudo-language element as a canonical description). The set of samples, enriched with grammatical rules, is integrated into the prompt to provide the LLM with a comprehensive linguistic and structural context. 
For spoken-to-sign, the LLM generates the corresponding translation in pseudo-language, which 
is then converted into the target language by the Sign Mapper, by mapping each part of the sequence to an entry from a predefined lexicon. The process is mirrored for the inverse sign-to-spoken translation task. 

\begin{figure}
\centering
\caption{Example of SignWriting (right), FSW encoding and corresponding descriptions of a sign language sequence (left). FSW provides a detailed, structured representation of signs, while sign descriptions offer a more abstract and language-independent representation.
} \label{fig:intro}
\vspace{-0.1in}
\includegraphics[width=.95\columnwidth]{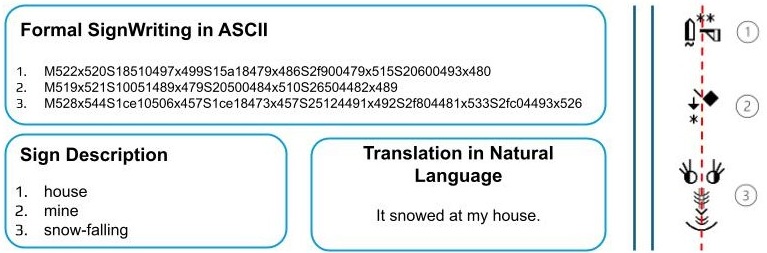}%\textwidth
\end{figure}

We evaluate our method on two datasets covering two different spoken and sign languages: spoken Italian to and from LIS (Italian Sign Language), and spoken English to and from ASL (American Sign Language). 
Although our method can be applied to any structured encoding of sign language, including parametric pose models such as SMPL-X, we evaluate our approach on a SignWriting computerized specification, namely Formal SignWriting (FSW), which provides a natural way to map predicted sequences with gold sequences, therefore simplifying evaluation. FSW represents signs sequences in a linearized and standardized format suitable for computational processing, which encodes signs using sequences of symbol identifiers and positional metadata, ensuring both syntactic rigor and spatial modeling capabilities (Figure~\ref{fig:intro}).
% an ASCII encoding of SignWriting, namely Formal SignWriting (FSW).
% While our method is general and can be extended to other structured representations, we validate it on 
This notation offers a compact, linguistically grounded, and implicitly explainable encoding of signs. It is widely adopted in education and in accessing sign language literature, and it is highly valued within the Deaf community.
%Furthermore, unlike recent approaches based on latent embeddings (e.g., Variational Autoencoders~\cite{van2017neural}), which require large-scale datasets and produce non-interpretable representations, FSW enables transparent, data-efficient translation pipelines, a particularly important feature in low-resource settings. \reminder{@misael: is it too much? yes :)}
Our experiments demonstrate that AulSign outperforms state-of-the-art models %for SignWriting translation 
in both spoken-to-sign and sign-to-spoken tasks, and show that LLM capabilities can be effectively leveraged for translation between spoken languages and underrepresented sign languages in a low-resource scenario. More importantly, this advancement has the potential to significantly enhance accessibility and inclusivity for the Deaf community.

%\vspace{0.05in}

\vspace{0.1in}
The contribution of this paper can be summarized as follows:
\begin{itemize}
    \item We present AulSign, a novel sign language translation model able to employ LLMs for translating from and to a language that have not been seen in their training process. Designed to operate in low-resource environments, AulSign addresses data scarcity by incorporating external lexicons and structured linguistic representations into an integrated pipeline.
    % \item Our method emphasizes explainability by mapping signs to a meta-lexicon, providing insights into the translation process. This feature enables transparency, allowing users to understand translation errors, identify potential misalignments, and trace the steps within the inference process.
    \item Our model maps signs to a meta-lexicon, offering explainability in the translation process. This allows users to understand translation errors, detect potential misalignments, and trace inference steps.
    \item 
    % By incorporating SignWriting as a structured, visually intuitive representation, AulSign enhances translation accuracy and expands usability. SignWriting’s detailed representation of handshapes, movements, facial expressions, and body positions streamlines the creation of visual content, such as animated avatars or videos, making spoken-to-sign translations more effective and user-friendly for Deaf audiences.
    % 
    AulSign improves translation accuracy and usability by using detailed intermediate representations like SignWriting and SMPL-X.
    % \reminder{Va a nostro svantaggio citare qui una rappresentazione linguistica che non stiamo usando? Si si va bene} 
    These help create visual content such as animated avatars, making spoken-to-sign translations more effective and accessible for Deaf users across various applications.
    % By incorporating structured and visually or semantically rich intermediate representations, AulSign enhances translation accuracy and expands usability. Representations such as SignWriting, which detail handshapes, movements, facial expressions, and body positions, or parametric models like SMPL-X, which encode full-body and facial pose dynamics, facilitate the generation of visual content such as animated avatars or videos. This flexibility makes spoken-to-sign translations more effective and user-friendly for Deaf audiences, enabling adaptation to a wide range of application scenarios, from textual rendering to realistic avatar animation. 
    % 
    \item We perform a comparative analysis of the performance of our method across different data availability scenarios and in a multi-lingual setting and show that AulSign
    % We outperform the state-of-the-art models for ASL and LIS in both spoken-to-sign and sign-to-spoken translation tasks. 
    outperforms state-of-the-art models in both spoken-to-sign and sign-to-spoken translation tasks.%, using FSW as the intermediate representation.
\end{itemize}

The paper is structured as follows:
% Section~\ref{sect:theoretical} outlines the theoretical background for sign language tasks. 
Section~\ref{sect:RW} reviews state-of-the-art methods for sign language translation.
% , focusing on spoken-to-sign and sign-to-spoken approaches. 
Section~\ref{sect:method} introduces the AulSign model, detailing its components and functionalities.
Section~\ref{sect:exp} describes the experimental background 
% (cf. Sect.~\ref{sect:theoretical}) 
and setup 
% (cf. Sect.~\ref{sect:text2sign} and Sect.~\ref{sect:sign2text})
and presents results for both translation directions in LIS and ASL. Section~\ref{sect:ablation} explores the impact of each AulSign's component, and Section~\ref{sect:discussion} analyzes the findings.
% , highlighting the model's significance and providing insights. 
% \reminder{lb: valutiamo se eliminate le limitations e nel caso mettiamole prima della conclusione. MM: ok lasciamole ma mettiamole priima della conclusione}
Finally, Section~\ref{sect:limit} presents the limitations of our work and Section~\ref{sect:conclusion} concludes the paper.

\begin{figure*}[ht]
\caption{Overview of the spoken-to-sign AulSign pipeline. We detail our method using the FSW notation as a reference.}
\includegraphics[width=\textwidth]{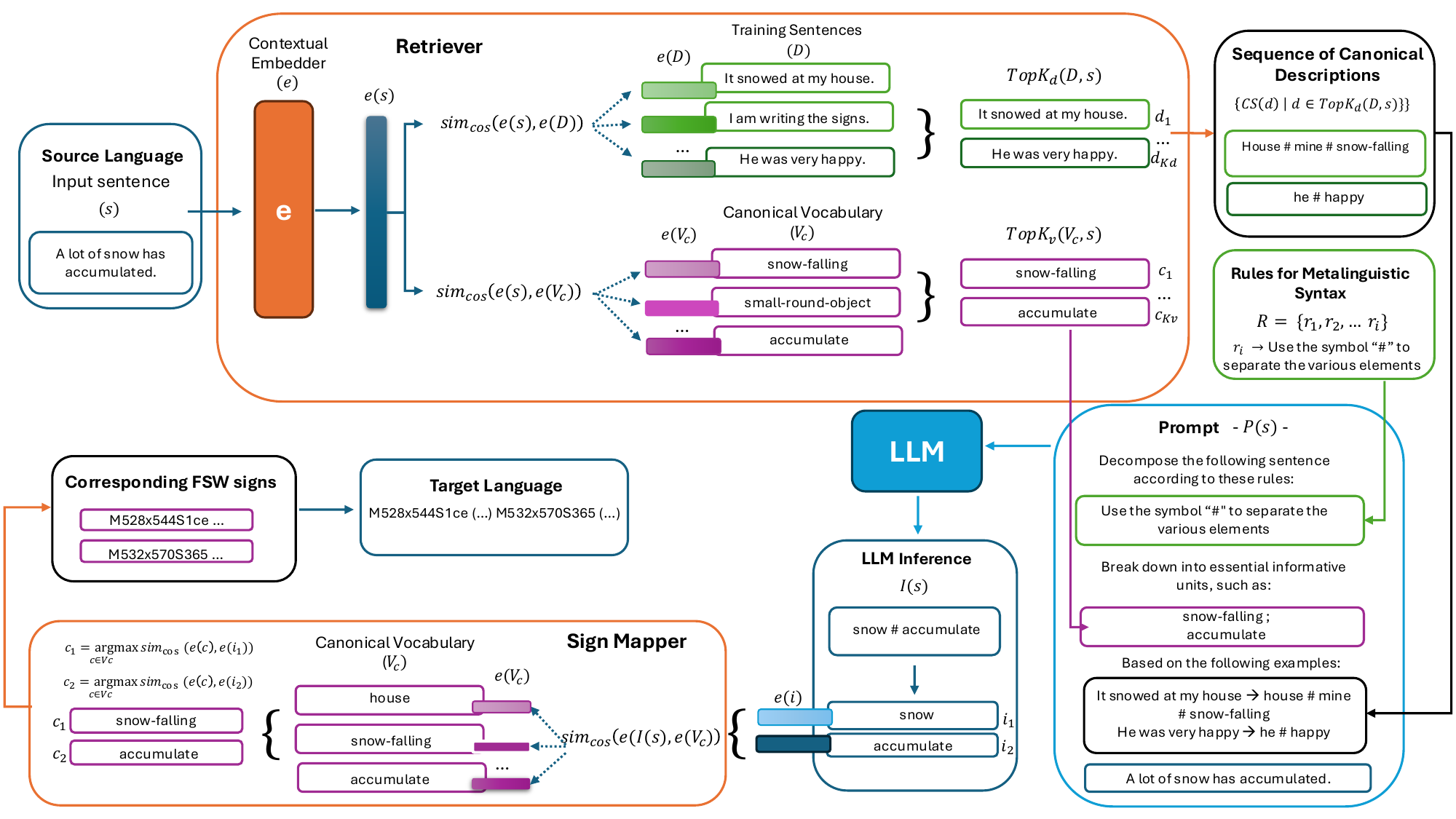}
 \label{fig1}
\end{figure*}

\section{Related Works}
\label{sect:RW}

Recent advancements in machine learning for SLT have primarily focused on two categories of models: end-to-end models, which directly map video inputs to text outputs, and representation-based models, which use intermediate representations (e.g., glosses or other meta-languages) to decompose the task into smaller, interpretable steps~\cite{de2024machine,farooq2021advances}. 
%lb + 
While end-to-end models are efficient~\cite{zhou2023gloss}, gloss-based models remain popular due to their interpretability and modularity. %However, generating sign language videos from text remains challenging, requiring temporal coherence and linguistic accuracy.
% While end-to-end models show efficiency and accuracy~\cite{zhou2023gloss}, their performance is not consistently superior to gloss-based models, which 
% remain a significant research focus due to their interpretability and modular architecture. Glosses, as symbolic representations of sign language, provide a structured mapping between signs and their corresponding linguistic or grammatical elements in natural language. This structured approach has shown particular effectiveness in unidirectional tasks, such as video-to-SL translation. 
Gloss-based multimodal models primarily address SL video-to-text translation. Inversely, SL text-to-video translation remains a more challenging task, requiring the generation of temporally coherent and linguistically accurate sign language videos. Therefore,
current systems for generating sign language from text rely mainly on animation or avatar-based methods~\cite{yu2025signavatars}.
Approaches based on an intermediate representation of sign language are helpful in this case, as they bypass the complexity involved in video generation and can be concatenated with a downstream avatar animation tool. %do not necessitate the processing of video input or the generation of video output. 
This study focuses on text-based translation systems that address both SL-to-text and text-to-SL tasks without relying on video data. State-of-the-art systems employ structured intermediate representations, including glosses~\cite{angelova2022using,yin2020better}, HamNoSys~\cite{kang2019spoken}, SignWriting and SMPL-X.
The integration of HamNoSys and glosses into a text-to-avatar pipeline presents limits since
HamNoSys lacks a standardized computational form suitable for direct integration into neural models, while glosses provide simple written approximations of signs using natural language labels. Gloss notation lacks the formalism to encode the phonological, spatial, and non-manual features intrinsic to sign languages, limiting their utility for synthesis or generation tasks. In contrast, representations like SignWriting or SMPL-X are explicitly designed to encode these visual-linguistic properties, making them more suitable for SL generation pipelines. %, including integration with parametric models such as SMPL-X. 
%
% Among these, SignWriting stands out due to its universal design and alignment with linguistic ontologies, which allow it to represent any sign language.
%
%Its potential integration into broader multimodal translation pipelines makes it a promising candidate for robust text-based SLT systems.
% Despite its simplicity, gloss-based representations tend to oversimplify the linguistic and spatial complexity of sign languages, losing crucial grammatical and iconic structures; therefore, we do not focus directly on glosses. 
Several studies have explored SignWriting’s recognition and segmentation~\cite{sevilla2023automatic,moryossef2023linguistically}, with most advancements focusing on languages like Arabic and Japanese~\cite{almasoud2011proposed,matsumoto2009jspad} and ontology-driven approaches~\cite{almasoud2012semsignwriting}. %, and multilingual scenarios~\cite{jiang2022machine,de2023leveraging}. 

Recent approaches employ transformers for translation~\cite{jiang2022machine,de2023leveraging}. Notably, the work of Jiang et al.~\cite{jiang2022machine} %and Freitas et al.~\cite{de2023leveraging}
provides key contributions to our research objectives.
%Jiang et al.~\cite{jiang2022machine} 
They propose a multilingual SLT system that translates natural language into SignWriting, setting the state-of-the-art for SignWriting-based SLT. Their multistep pipeline involves analyzing, factorizing, decoding, and evaluating SignWriting sequences to generate natural language text. The system employs sequences of FSW, 
% , a computationally standardized representation of SignWriting symbols (cf. Figure~\ref{fig:intro}).
% FSW sequences 
which are decomposed into base units (symbols) and supplementary information (e.g., positional numbers). These elements are encoded as variables
processed by a factorized model~\cite{koehn2007factored,garcia2016factored}. For spoken-to-sign tasks, symbol tokens are decoded using Beam Search, while positional factors are predicted through regression. Additionally, dictionary data (e.g., Dictionary Puddles)\footnote{https://www.signbank.org/signpuddle/} are integrated to enhance model performance. Despite these advancements, the model shows some limitations. For example, finger-spelling requires specialized handling, and traditional text classification metrics may be insufficient for evaluating SLT due to its visual-spatial characteristics.
Recent advancements have leveraged LLMs to enhance SLT, particularly in video-to-sign language tasks~\cite{gong2024llms,hwang2024efficient,lim2023sign,wong2024sign2gpt,jiang2024signclip}. These methods have been explored with and without intermediate representations like gloss notation~\cite{liu2024scope,johnson2024innovative}. Furthermore, LLMs have demonstrated potential in translating between text and glosses~\cite{fayyazsanavi2024gloss2text,lee2023leveraging}. However, to the best of our knowledge, no research has directly addressed the task of Sign Language translation using generative pretrained LLMs from natural language to encoded language representations, such as SignWriting. This aspect poses challenges since
LLMs have not seen sign language representations during their training phase.

\vspace{-0.1in}

\section{AulSign: A novel LLM-based method for Sign Language Translation}
\label{sect:method}

Our method comprises three main components: a Retriever, an LLM, and a Sign Mapper. The Retriever 
identifies samples from the training set\footnote{we refer to the set of available sentences with corresponding translation, although our method do not require explicit training}
% \reminder{lb: modifichiamo in dataset? training set è un po' misleading a detta di un revisore. MM: OK ho messo una footnote} to be included in the prompt to instruct the LLM. 
% \reminder{gt ansewer: secondo me training set è da evitare assolutamnte, perchè il rev. si confonde 100/100. Dataset potrebbe andar bene, ma non è troppo generico? Perchè ho paura che potrebbe pensae che abbiamo barato. Io metterei:  from the knowledge base oppure from the reference set ed eventualmente pois gli si spiega se ci solleva la domanda}
% 
The training set is pre-processed offline to convert signs into canonical descriptions, which serve as pseudo-language expressions capturing the meaning of each sign.
Compared to glosses, canonical descriptions are longer, more expressive, and designed to be unique for each specific meaning, allowing a one-to-one correspondence between a description and a sign. This design avoids the ambiguity arising from polysemy and enhances semantic precision.
By maintaining explicit and interpretable mappings, our method not only improves explainability but also allows adaptation to different structured sign encodings, such as SignWriting, HamNoSys, and SMPL-X.

Each sample of the pre-processed training set contains the spoken text and a corresponding decomposition into a sequence of canonical descriptions. 
Samples that are returned by the Retriever are used for prompt generation.
This task combines the retrieved sample sentences, paired with their corresponding decompositions,
with samples from 
a structured vocabulary and a set of predefined grammar rules to employ a structured prompt, which enables the LLM to perform few-shot inference on the input sentence.
The Sign Mapper is employed only in spoken-to-sign translation and converts the sequence of canonical descriptions generated by the LLM into a sequence of signs.
The whole process is described in Figure~\ref{fig1}. Next we detail the various parts of our pipeline in the spoken-to-sign direction. The inverse process is similar, with small adjustments, which are discussed at the end of the section.

Our method considers a training set $D$ of spoken sentences associated with their sign's encoding counterpart, i.e. sequences of signs, and a vocabulary of signs $V_s$, where each sign is associated with one or more descriptions in natural language. We pre-process both $D$ and $V_s$ to translate the encoded sign sequences into sequences of canonical descriptions\footnote{for the Italian LaCAM CNR-ISTC dataset we do not need this step since in the corpus all signs have been manually associated to canonical descriptions}. First, we define an equivalence operator\footnote{in our implementation signs are considered equivalent if they contain the same set of symbols with associated orientation and rotation; more complex notions that also consider the spatial position can be employed to improve results} $\equiv$ between signs and merge equivalent signs into one single entry, choosing the most frequent sign as a representative. Then, we construct a set $V_c$ of unique and non-ambiguous canonical descriptions, by choosing among all descriptions associated to a sign the most frequent ones and combining them into a single string. 
Finally we substitute the sequences of signs associated to sentences in $D$ with corresponding sequences of canonical descriptions by probing, for each sign of the sequence, an equivalent sign from $V_s$ and taking the corresponding canonical description. If an equivalent sign exists in $V_s$, it is unique because all equivalent signs have been previously merged in $V_s$. On the other hand, if an equivalent sign is not found, a special ``<unk>'' description is considered. Given a spoken sentence $d \in D$, we denote as $CS(d)$ its associated sequence of canonical descriptions\footnote{further details and examples are provided in the supplemental material}.

The Retriever identifies the top-\(K\) sentences from the training set $D$ that are most semantically similar to the given input sentence \(s\). This process ensures that the subsequent prompt generation 
operates with contextually relevant examples.
To improve results, the Retriever also identifies a set of canonical descriptions from $V_c$ that are similar to the input sentence, to be fed to the LLM as samples of canonical descriptions.
Both retrieval steps 
employ a contextual embedder \(e(\cdot)\), trained for semantic similarity, to encode both the input sentence and the candidate sentences 
into high-dimensional representations. The similarity between \(s\) and each candidate 
is computed by cosine similarity.
The system selects the  top-\(K\) most semantically similar entries, %forming 
which we denote with 
the sets \emph{TopK\textsubscript{d}}\((D,s)\) and \emph{TopK\textsubscript{v}}\((V_c,s)\), respectively.

The prompt generation step develops an input-sentence-customized prompt, leveraging on the few-shot learning abilities of LLMs. 
We design the prompt 
by combining grammar rules (\(R=\{r_{1},r_3,\ldots r_{i}\}\)) with illustrative examples -- retrieved sentences with their canonical descriptions -- and the input sentence. 
The retrieved canonical descriptions from the Retriever are also included in the prompt as examples of canonical descriptions that the LLM can use to break down the input sentence. 
The complete prompt structure is shown in Figure \ref{fig1}.

The LLM generates a sequence $I(s)$ of strings in a form that mimics canonical descriptions in $V_c$, which feeds the Sign Mapper. To match the generated pseudo-canonical descriptions to entries in $V_c$ we again use semantic similarity. We employ the same contextual embedder $e(\cdot)$ used for retrieval and, for each pseudo-canonical $i_l \in I(s)$, we extract:
$c_l = \arg\max_{c \in V_c} \text{ 
 sim}_{\text{cos}}(e(c),e(i_l))$.
Finally, the model retrieves the corresponding encoded signs associated with the selected canonical descriptions to generate the final translation. 
% Note that the LLM always produces signs that are in $V_c$ and it cannnot generate UNK sign as our method is based on this constrain. 
%
The output of AulSign can be provided to an avatar animation system, which can be implemented rule based, i.e. associating real poses and movement to SignWriting glyphs, or by interpolating between key SMPL-X poses~\cite{aziz2023evolution}.

For the sign-to-spoken task, the input sentence in sign language notation is first converted into a sequence of canonical descriptions in the same way as for the training set pre-processing. Signs of the input sequence are probed from $V_s$ using the equivalence operator $\equiv$ and the corresponding canonical descriptions are taken. Similar sequences of canonical descriptions are then extracted from $D$ and used with their speech counterpart to instruct the LLM, i.e. to build the prompt with few-shot samples. The prompt is built in the same way as for speech-to-sign, except that the examples for the break down are not given, since they are not necessary. The output of the LLM is then returned as the resulting spoken sentence.
% \reminder{lb: ho eliminate i riferimenti al FSW nel metodo parlando di encoded sign or sign language notation. MM: ottimo}

\section{Experiments and Results}
\label{sect:exp}

To assess the effectiveness of AulSign, we conduct experiments on two benchmark datasets: the English SignBank+~\cite{moryossef2023signbank+} for ASL and the LaCAM CNR-ISTC Italian dataset for LIS~\cite{olga_dataset25}.
% an Italian dataset based on SignPuddle\footnote{https://www.signbank.org/signpuddle/} for LIS.
% \reminder{gt:anche qui vale lo stesso discorso dell'attacabilità per lis fatto nell'intro? LB: no, qui non dovrebbe valere dato che stiamo specificando più volte il perchè usiamo il Sw come caso studio per validate anche se il sistema supports altri sistemi di notazione.. cmq rappresenta il maggior pt a sfavore dell'intero paper second me  ma se ci attaccano su questo possiamo fare ben poco (l'alternativa sarebbe presentare i risultati su più di una notazione - ad esempio anche su SMP.X - ma a quel punto non ne usciamo più).}
SignBank+ is a multilingual corpus with 254,002 distinct elements, 76 sign language encodings, and 153 ``puddles'', systems that categorize signs by language or dialect. The dataset is organized into three subcorpora: Manually (cleaned via manual review), GPT-3.5 (cleaned using GPT-3.5 with a few-shot learning paradigm~\cite{brown2020language}), and Bible (aligned biblical texts). To ensure consistency and reduce noise, we focus exclusively on the English subset for our experiments.
The English subcorpus of SignBank+ comprises 43,705 annotated items spanning nine variants of English Sign Language, including 19,304 unique signs and 13,631 sentences. We use the unique-sign English SignBank+ subcorpus to build \(V_c\) (cf. Sect.~\ref{sect:method}) and split the sentence-based subcorpus into training set $D$ (cf. Sect.~\ref{sect:method}) and test set. We experiment with three different training set configurations (referred to as I, II, III) to assess the method's robustness across varying data conditions. The first configuration considers the entire training set, %, encompassing both In-Vocabulary (i.e. signs present in the Vocabulary) and Out-Of-Vocabulary (i.e. signs not present in the Vocabulary) signs. This configuration, 
which counts 13,275 sentences.
%, tests the model’s ability to generalize in a data-rich setting with mixed data quality.
Note that not all signs in the sentences are covered by our vocabulary (extracted by single sign sentences), therefore our method, which requires a complete vocabulary, is at a disadvantage in this setting\footnote{we identify the extraction of vocabulary from sentences as an interesting research direction}. 
% By incorporating 13,275 sentences into the training set, this approach allows a comprehensive evaluation of the model's performance in a data-rich scenario, focusing on its ability to handle signs lacking direct dictionary equivalents.
The second configuration filters out sentences containing out-of-vocabulary signs, resulting in a more consistent dataset of 2,301 sentences. The third configuration simulates a low-resource scenario by randomly sampling 115 sentences from the training set of the second configuration. 
 Notably, although the number of training samples varies across configurations, the vocabulary \(V_c\) remains fixed as it serves as a connection between canonical descriptions and their corresponding entries in the encoded sign representation. 
For comparisons, we train the state-of-the-art model from Jiang et al.~\cite{jiang2022machine} using three parallel configurations (i.e. I, II, and III). Each model is trained on the same set 
% \reminder{è lo stesso insime di frasi giusto? ho sotituito number con set} 
of training sentences as the corresponding AulSign configurations and includes the complete vocabulary (i.e. \(V_c)\) in each configuration training process. Following Jiang et al., we configure the set of hyperparameters with a learning rate of 0.0001, a drop-out of 0.50, 
% \reminder{non è troppo alto?}
a batch size of 64\footnote{in place of 32 used by Jiang et al. to improve efficiency} and a learning-rate-factor of 0.70. 
% \reminder{bisogna specificare come i parametri sono stati scelti. Sono quelli che usano loro nel loro paper?} 
We evaluate both approaches on a test set of 356 sentences.

% an Italian dataset based on SignPuddle\footnote{https://www.signbank.org/signpuddle/} for LIS.
The LaCAM CNR-ISTC corpus~\cite{olga_dataset25} is an Italian dataset based on SignPuddle\footnote{https://www.signbank.org/signpuddle/} for LIS. It contains 2,149 annotated items, including 1,974 signs and 782 sentences. Each sign is associated with a hand-made canonical description and a FSW representation. We use signs to populate the vocabulary \(V_c\) and split the dataset into training and test sets of 547 and 235 sentences, respectively.
 Due to the limited data available in the Italian LIS dataset, we do not define three separate configurations as in the ASL task. 

To retrieve relevant examples for the AulSign Retriever module, we employ the Lee et al.~\cite{emb2024mxbai} and the Reimers et al.~\cite{reimers-2019-sentence-bert} models for ASL and LIS, respectively, as embedding models. We set the number of top retrieved sentences \(K_d\) to 20 and the number retrieved canonical descriptions \(K_v\) to 100.

For prompt generation (cf. Sect.~\ref{sect:method}), we define two distinct sets of rules to structure the metalinguistic syntax settings. These rules describe each sign based on its canonical description and are supplemented with illustrative examples of sentences and their associated canonical descriptions to ensure clarity and alignment between linguistic concepts and their representations. We define a total of seven rules for both English and Italian\footnote{for an overview of the grammar rules employed, we refer to \url{https://github.com/gabrix00/AulSign}}. 
For all experiments, we employ GPT-3.5 as an LLM.\footnote{https://platform.openai.com/docs/models/gpt-3-5-turbo}

Section~\ref{sect:text2sign} details the overall performance of the AulSign method for LIS and ASL in executing the spoken-to-sign translation task. Section~\ref{sect:sign2text} presents the results for the inverse task, i.e., sign-to-spoken translation. To provide a deeper understanding of our results, we performed both an error analysis and a cost analysis, the details of which are presented in the supplemental materials.

\subsection{Spoken-to-sign translation} 
\label{sect:text2sign}

Table~\ref{tab:english} presents the overall performance of the AulSign method on the English SignBank+ dataset in terms of F1-score, BLEU~\cite{papineni2002bleu}, ChrF2~\cite{popovic2015chrf}, and Mean Absolute Error (MAE). We consider the F1-score as an appropriate metric to evaluate predictions at the symbol (with associated rotation and orientation) level,
since the order of symbols within a sign is not semantically relevant\footnote{for further information about the structure of FSW we refer to~\cite{jiang2022machine}}. 
 We compute the F1 score on a sentence-by-sentence basis by considering all signs collectively and calculating the harmonic mean of precision and recall. We then report the average across the entire dataset. This approach accounts for the challenge of aligning gold and predicted signs within a sentence, especially when their lengths or ordering differ.
However, symbol-level evaluation alone is insufficient for FSW, due to the semantically relevant order across signs. 
To address this limitation, and following Jiang et al.~\cite{jiang2022machine}, we also employ the popular translation metrics BLEU~\cite{papineni2002bleu} and ChrF2~\cite{popovic2015chrf},
which capture statistics at both word and character levels. Additionally, we assess positional values (x and y) using MAE to quantify the discrepancy between predicted and ground-truth values. As the symbol order within a sign is not semantically significant, both gold standard and predicted sequences are alphabetically sorted before evaluation.

\begin{table}
  \centering
\resizebox{\linewidth}{!}{
\begin{tabular}{@{}clcccccc@{}}
\toprule
\textbf{Setting} & \textbf{Model} & \textbf{F1} & \textbf{BLEU}  & \textbf{ChrF2} & \textbf{MAE X} & \textbf{MAE Y} & \textbf{Training size} \\ 
% & & & & & & \\
\midrule
I & Jiang et al.                     & \textbf{0.45}       & \textbf{29.26} & \textbf{57.82} & \textbf{23.77} & \textbf{27.96} & \multirow{2}{*}{13,275 + $|V_c|$} \\
I & AulSign                           & 0.42                & 25.40           & 56.44          & 25.02          & 29.66          &                        \\ \midrule
II & Jiang et al.                    & 0.40                 & 22.68          & 50.65          & \textbf{23.78} & \textbf{27.80}  & \multirow{2}{*}{2,301 + $|V_c|$}  \\
II & AulSign                         & \textbf{0.41}       & \textbf{23.96} & \textbf{56.06} & 25.42          & 30.31          &                        \\ \midrule
III & Jiang et al.                   & 0.27                & 10.94          & 39.17          & \textbf{23.81} & \textbf{27.82} & \multirow{2}{*}{115 + $|V_c|$}   \\
III & AulSign                        & \textbf{0.37} (±0.006)      & \textbf{18.79} (±0.665)   & \textbf{53.91} (±0.153) & 25.52 (±0.138)         & 30.95 (±0.188)         &                        \\ \bottomrule
\end{tabular}}
\caption{Overall spoken-to-sign ASL translation models performance on the English SignBank+ dataset, in terms of F1 score, BLEU, ChrF2, and MAE. We test three AulSign configurations: (I) data-rich (13,275 sentences), (II) filtered (2,301 sentences), and (III) low-resource (115 sentences). $|V_c|$ refers to the totality of vocabulary entries. We compare the results to the baseline model (Jiang et al. \cite{jiang2022machine}) under the same conditions, with bold values indicating the best result per configuration.} \label{tab:english}

\end{table}

\vspace{-0.1in}

\begin{figure}[h]
% \begin{minipage}[t]{0.3\textwidth}
\centering
\caption{
%\reminder{il testo nella figura deve essere più grande}
Performance trend of AulSign and the state-of-the-art model across different training set sizes, evaluated in terms of F1-score (first column Table~\ref{tab:english}) for the ASL spoken-to-sign translation task.} \label{fig:f1}
\includegraphics[width=0.48\textwidth]{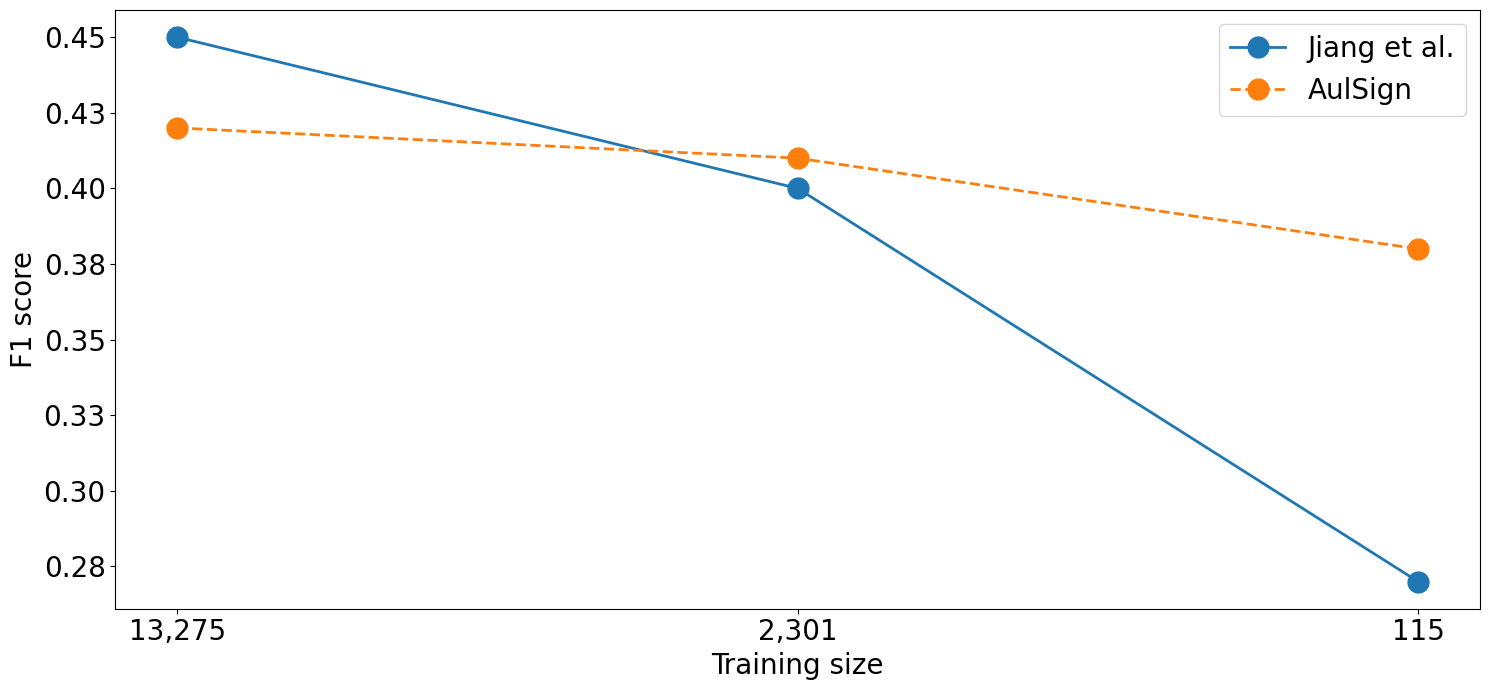}
% \end{minipage}%
\end{figure}

As shown in Table~\ref{tab:english}, AulSign consistently outperforms the Jiang et al. baseline in data-scarce scenarios. 
In this scenario, the model achieves an F1 score of 0.37, a 10-point improvement over Jiang et al. III configuration (i.e. 0.27). This trend is mirrored in BLEU and ChrF2 scores, where the third AulSign configuration attains 18.79 and 53.91, compared to 10.94 and 39.17 for the baseline. These results underscore the AulSign’s ability to handle low-resource settings.
In data-rich settings (I and II), AulSign demonstrates comparable performance to Jiang et al., even when confronted with noisy data. AulSign I achieves an F1 score of 0.42, slightly lower than Jiang et al. I (i.e. 0.45), while maintaining competitive BLEU (i.e. 25.40 vs. 29.26) and ChrF2 scores (i.e. 56.44 vs. 57.82). Our assessment affirms AulSign's robustness across different data conditions.
Regarding MAE, Jiang et al.’s models generally achieve lower positional errors, with minimum values (in configuration II) of 23.77 and 27.80 for X and Y coordinates, respectively. AulSign’s MAE values are slightly higher (25.42 and 30.31 for the same configuration), reflecting a trade-off between precise factor prediction and overall symbol recognition quality. This discrepancy can be attributed to the architectural focus of Jiang et al.’s models on positional accuracy, whereas AulSign employs a general-purpose translation framework. Nevertheless, AulSign achieves competitive overall performance, highlighting its efficacy in handling both sequence-level and positional challenges in SignWriting-based translation.
To better illustrate the performance trend as the amount of training data varies, we report in Figure~\ref{fig:f1} the F1-score column of Table~\ref{tab:english}.  AulSign shows consistent performance across training set sizes, contrasting with Jiang et al.’s model, which exhibits significant degradation in low-resource conditions. %This highlights AulSign’s superior generalization and adaptability to diverse data scenarios.
To assess the robustness of our method even with a small number of samples, in the low-resource configuration (III) we compute the average and standard deviation of 10 runs, each of them obtained with a different subset of 115 sentences. In the other configurations we use the whole set of available sentences (with and without ``<unk>''), therefore we cannot perform this test.
Results show a standard deviation of less than 1\%.

\vspace{0.1in}

Table~\ref{tab:italian} summarizes the performance of AulSign on the Italian LaCAM CNR-ISTC corpus, evaluated using F1-score, BLEU, ChrF2, and MAE for positional coordinates. 
 Since the Italian setting represents a low-resource scenario with limited data availability, we cannot compute standard deviations or report results for additional data configurations, as further subdivisions would be uninformative. 
AulSign achieves substantial improvements across all metrics compared to Jiang et al. Notably, it attains an F1-score of 0.63, representing a 13-point increase over the baseline (0.50), demonstrating superior capability in sign classification and identification.
For sequence-level metrics, AulSign outperforms Jiang et al. with BLEU and ChrF2 scores of 37.71 and 57.54, respectively, compared to the baseline’s 16.40 and 45.18. Additionally, AulSign demonstrates superior positional accuracy, with MAE values of 21.22 (X) and 33.22 (Y), compared to 23.82 and 37.44 for Jiang et al. This indicates AulSign’s ability to precisely model even the spatial 
%and temporal 
characteristics of signs %, critical for effective SignWriting representation.
when a comprehensive vocabulary is available.

\begin{table}
  \centering
%\resizebox{\linewidth}{!}
{
\begin{tabular}{lccccc}
\toprule
\textbf{Model} & \textbf{F1} & \textbf{BLEU} & \textbf{ChrF2} & \textbf{MAE X} & \textbf{MAE Y} \\
\hline

Jiang et al.                       & 0.50                            & 16.40                              & 45.18                               & 23.82                              & 37.44                              \\
AulSign                            & \textbf{0.63}                   & \textbf{37.71}                     & \textbf{57.54}                      & \textbf{21.22}                     & \textbf{33.22}                     \\ \bottomrule 
\end{tabular} }
\caption{Models overall performance in spoken-to-sign Italian-to-LIS translation using FSW-encoded sequences on the Italian LaCAM CNR-ISTC corpus.
}\label{tab:italian}
\end{table}

\vspace{-0.3in}

\subsection{Sign-to-spoken translation}
\label{sect:sign2text}

Tables~\ref{tab:en_vice} and~\ref{tab:it_vice} show the results on sign-to-spoken translation for ASL-to-English and LIS-to-Italian, using FSW-encoded sign language sequences.
% We evaluate model performance in terms of BLEU and ChrF2, 
% BLEU assesses sequence-level precision, while ChrF2 evaluates character-level similarity, including morphological context.
% We conduct experiments on ASL 
As for spoken-to-sign,
%To assess the impact of different training strategies, 
we evaluate three AulSign model configurations (I, II, III) for ASL and one for LIS in terms of BLEU and ChrF2. We do not compute F1 and MAE since the former is not adequate to assess spoken natural language sentences and the latter has no meaning in this context. 
For a comprehensive overview of model configurations and experimental setup, we refer to the beginning of this section. %Section~\ref{sect:exp} and Section~\ref{sect:text2sign}.
% Both ASL and LIS experiments adhere to the experimental setup detailed in Section~\ref{sect:exp}. 
% For detailed prompt configurations,  consult the GitHub repository.~\footnote{}\reminder{inserisci}

% As detailed in Section~\ref{sect:text2sign}, we evaluate the performance of three AulSign model configurations (I, II, III) in terms of BLEU and ChrF2 to assess the impact of different training strategies. Due to limited training data availability for LIS, we employ a single configuration for the Italian language. Both ASL and LIS experiments use training data configurations and vocabularies aligned with those used for English and Italian, as detailed in Section~\ref{sect:exp}. For detailed prompt configurations, refer to the supplementary documentation on our GitHub repository\footnote{}\reminder{inserire}.

As shown in Table~\ref{tab:en_vice} and Table~\ref{tab:it_vice}, AulSign consistently outperforms the baseline, achieving state-of-the-art results for both ASL and LIS. In a low-resource setting for the ASL, AulSign significantly outperforms the Jiang et al. model, yielding a BLEU score of 26.59 and a ChrF2 score of 40.76, compared to the baseline's 2.20 and 18.90, respectively.
Similarly, AulSign achieves a BLEU score of 17.95 and a ChrF2 score of 50.42, significantly outperforming the baseline scores of 6.50 and 29.40 in the low-resource LIS setting.
% in the low-resource setting of LIS, AulSign exhibits a significant performance improvement over the baseline model, achieving a BLEU score of 17.95 and a ChrF2 score of 50.42, compared to baseline scores of 6.50 and 29.40.
In high-resource settings, we evaluate two configurations of AulSign (i.e. I and II). Both configurations show superior performance compared to the Jiang et al. model.  
Specifically, the first configuration of AulSign achieves a BLEU score of 23.38 and a ChrF2 score of 39.21, exceeding the baseline scores of 18.40 and 34.40. Notably, the second configuration, trained on a reduced dataset, significantly outperforms the baseline, with a BLEU score of 24.75 and a ChrF2 score of 40.26, compared to the baseline's 8.40 and 22.50, respectively. In low-resource scenarios, the low performances of Jiang et al. are expected since there are not sufficient data for training, considering the complexity of natural language. In any case, our model can leverage the language processing abilities of pretrained LLMs, therefore producing higher-quality translation.
 As for the spoken-to-sign task (cf. Sect.~\ref{sect:text2sign}), we also computed an average on 10 different runs to assess the robustness of our method for the ASL low-resource scenario (configuration III, not reported). Performance resulted similar to those of Table~\ref{tab:en_vice}, with a standard deviation of less than 1\%.  
%Furthermore, the AulSign model exhibits superior robustness to variations in training data size compared to the Jiang et al. model, which demonstrates a significant performance degradation with reduced training data.

\begin{table}
  \centering
%\resizebox{0.9\linewidth}{!}
{
\begin{tabular}{@{}clccc@{}}
\toprule
\textbf{Setting} & \textbf{Model} & \textbf{BLEU} & \textbf{ChrF2}  & \textbf{Training Size} \\ \hline
I & Jiang et al.   & 18.40 & 34.40 & \multirow{2}{*}{13,275 + $|V_c|$} \\
I & AulSign        & \textbf{23.38} & \textbf{39.21} \\ \midrule
II &  Jiang et al.  & 8.40 & 22.50 & \multirow{2}{*}{2,301 + $|V_c|$}\\
II & AulSign       & \textbf{24.75} & \textbf{40.26}  \\ \midrule
III & Jiang et al. & 2.20 & 18.90 & \multirow{2}{*}{115 + $|V_c|$}\\
III & AulSign & \textbf{26.59} & \textbf{40.76} \\ \bottomrule
\end{tabular} }
\caption{Overall sign-to-spoken ASL translation models performance on the English SignBank+ dataset, in terms of BLEU, and ChrF2. We test three AulSign configurations: (I) data-rich (13,275 sentences), (II) filtered (2,301 sentences), and (III) low-resource (115 sentences). $|V_c|$ refers to the totality of vocabulary entries. We compare the results to the baseline model (Jiang et al. \cite{jiang2022machine}) under the same conditions. Bold values indicate the best result per configuration.}
% \caption{Models overall performance in sign-to-spoken ASL translation on the English SignBank+ dataset, evaluated in terms of BLEU and ChrF2. Three AulSign configurations are tested: (I) data-rich with 13,275 sentences, (II) filtered with 2,301 sentences excluding Out-Of-Vocabulary signs, and (III) low-resource with 115 randomly sampled sentences. Results are compared to a baseline model under identical conditions, with bold values indicating the best results per configuration.}
\label{tab:en_vice}
\end{table}

\vspace{-0.2in}

\begin{table}
  \centering
%  \small
% \resizebox{\linewidth}{!}{
% \begin{minipage}{0.45\textwidth}
\begin{tabular}{lcc}
\toprule
\textbf{Model} & \textbf{BLEU} & \textbf{ChrF2} \\ \midrule
Jiang et al.   & 6.50 & 29.40 \\
AulSign        & \textbf{17.95} & \textbf{50.42} \\ \bottomrule
\end{tabular}
\caption{Overall performance in sign-to-spoken LIS translation using FSW-encoded sequences on the Italian LaCAM CNR-ISTC dataset, evaluated in terms of BLEU and ChrF2.}
\label{tab:it_vice}

\end{table}

\vspace{-0.35in}

\section{Ablation Study}
\label{sect:ablation}

%\reminder{non dovrebbe essere una subsection? LB: se vogliamo si.. ma nei paper non è solitamente inserita come section?}
\begin{table*}[h]
  \centering
%\resizebox{\linewidth}{!}
{
\begin{tabular}{llccccc}
\toprule
\textbf{Task} & \textbf{Embedder} & \textbf{F1} & \textbf{BLEU} & \textbf{ChrF2} & \textbf{MAE X} & \textbf{MAE Y} \\
\hline
Text-to-Sign & MPnet & 0.34 (± 0.005) & 14.71 (± 0.444) & 54.38 (± 0.174) &	25.13 (± 0.111) 	& 31.25 (± 0.191) \\ 
Text-to-Sign & MiniLM & 0.34 (± 0.00)       & 15.16 (± 0.425)  & 54.73 (± 0.248) & 25.71 (± 0.137) & 31.05 (± 0.122)           \\ 
Text-to-Sign & Mxbai & 0.34 (± 0.004) & 15.43 (± 0.316) & 54.92 (± 0.213) &	25.36 (± 0.126) & 31.36 (± 0.259) \\ \hline
Sign-to-Text & MPnet & - &  19.92 (± 0.638) & 39.91 (± 0.230) & - & - \\
Sign-to-Text & MiniLM & - & 19.05 (± 1.590) & 39.55 (± 1.063) & - & - \\
Sign-to-Text & Mxbai & - & 18.89 (± 1.909) & 39.13 (± 1.046) & - & - \\
\bottomrule
\end{tabular} }
\caption{Spoken-to-Sign and Sign-to-Spoken ASL ablation study of AulSign focusing on the embedder.%, in terms of F1 score, BLEU, ChrF2, and MAE.
% \reminder{@misael, va bene dettagliare così la teb? ablation study on ...? o è meglio usare un altro format? MM: sistemato. MM: invece, se c'è spazio mettiamo queste due tabelle in doppia colonna (testo troppo piccolo) con begin{table*} e end{table*}} 
We test three different embedder models (i.e. MPnet, MiniLM and Mxbai) on the English SignBank+ dataset in a low-resource scenario (i.e. 115 items). For each score, we report the mean of 10 runs with the standard deviation.   
% al variare degli embedder EN - text to sign
}\label{tab:emb en 1}
\end{table*}

%Risultati dell'ablation with LLMs.
\begin{table*}[h]
  \centering
%\resizebox{\linewidth}{!}
{
\begin{tabular}{clccccc}
\toprule
\textbf{Task} & \textbf{LLM} & \textbf{F1} & \textbf{BLEU} & \textbf{ChrF2} & \textbf{MAE X} & \textbf{MAE Y} \\
\hline
Text-to-Sign & LlaMa-70B & 0.34 (± 0.004) & 15.43 (± 0.316) & 54.92 (± 0.213) &	25.36 (± 0.126) 	& 31.36 (± 0.259) \\ 
Text-to-Sign &  GPT-3.5 & 0.37 (± 0.006)       & 19.73 (± 0.735)  & 55.07 (± 0.159) & 25.52 (± 0.138) & 30.95 (± 0.188)           \\ 
Text-to-Sign & GPT-4.1-nano & 0.35 (± 0.003) & 17.66 (± 0.145) & 55.52 (± 0.196) &	25.18 (± 0.077) & 30.77 (± 0.147) \\ \hline
Sign-to-Text & LlaMa-70B & - & 18.89 (± 1.909) & 39.14 (± 1.046) & -& -\\ 
Sign-to-Text & GPT-3.5 & - & 24.22 (± 1.139)  & 40.04 (± 0.750) & -& - \\ 
Sign-to-Text & GPT-4.1-nano & - & 15.29 (± 0.718)	& 32.38 (± 0.652) & -& -\\
\bottomrule
\end{tabular} }
\caption{Spoken-to-Sign and Sign-to-Spoken ASL ablation study focusing on the LLM.% in terms of F1 score, BLEU, ChrF2, and MAE. 
We test AulSign performance at the variation of three different LLMs (i.e. LlaMa-70B, GPT-3.5 and GPT-4.1-nano) on the English SignBank+ dataset in a low-resource scenario (i.e. 115 items). For each score, we report the mean of 10 runs with the standard deviation. }\label{tab:ablation llm}
\end{table*}

We assess the contribution of the AulSign framework's individual components by analyzing the impact of its three core modules: the Retriever, the LLM, and the Sign Mapper (cf. Sect.~\ref{sect:method}). 
The ablations are conducted on both spoken-to-sign and sign-to-spoken tasks for ASL\footnote{we report ablation studies for LIS in the supplementary materials, as they have similar performance to those of ASL}. To assess AulSign robustness, we experiment models performance on a low-resource setting, i.e., on the 115 items of configuration III (cf. Sect.~\ref{sect:sign2text}), reporting the mean over 10 different runs with the corresponding standard deviation. 

Table~\ref{tab:emb en 1} examines how variations in the embedding model affect performance, highlighting the roles of the Retriever and the Sign Mapper.
The embedder plays a critical role in two stages: (i) during the Retriever phase, where it selects the most semantically similar sentences from the training set to construct the LLM prompt, and (ii) in the Sign Mapper module, where it aligns the pseudo-canonical descriptions generated by the LLM with entries from the canonical vocabulary. We report the performance of three well-known embedding models, i.e., %MPnet\footnote{https://huggingface.co/sentence-transformers/all-mpnet-base-v2}, MiniLM\footnote{https://huggingface.co/sentence-transformers/all-MiniLM-L6-v2} and Mxbai\footnote{https://huggingface.co/mixedbread-ai/mxbai-embed-large-v1}, \reminder{@gabriele, esistono dei paper legati a quest modelli di embedding? se si, potresti sostituire il link con il riferimento bib?}
MPnet \cite{reimers-2020-multilingual-sentence-bert}, MiniLM \cite{reimers-2020-multilingual-sentence-bert} and Mxbai \cite{emb2024mxbai} in terms of F1-score, BLEU, and ChrF2. The MPNet model, based on a masked and permuted pre-training architecture, shows strong performance in capturing semantic similarity and provides high accuracy across standard benchmarks. The MiniLM model, while lighter and faster due to its distilled transformer architecture, offers a good trade-off between performance and efficiency, making it suitable for real-time or resource-constrained environments. Finally, the Mxbai model, a large-scale embedder optimized through instruction tuning, outperforms the others in terms of generalization and retrieval tasks, demonstrating particularly robust results in multi-domain and cross-task scenarios.
As shown in Table~\ref{tab:emb en 1}, all models exhibit comparable performance, with low standard deviation (i.e., less than 2\%) across both text-to-sign and sign-to-text tasks. Specifically, MxBai achieves slightly higher scores in the text-to-sign direction, while MPNet performs marginally better in the sign-to-text task. However, these differences are minimal, suggesting AulSign is robust to changing the embedder.

Table~\ref{tab:ablation llm} analyzes the impact of the LLM module by comparing three different AulSign's configurations based on distinct LLMs. Specifically, we evaluate the performance of LLaMA-3 70B\footnote{https://ollama.com/library/llama3.3:70b-instruct-fp16}, GPT-4.1-nano\footnote{https://platform.openai.com/docs/models/gpt-4.1-nano}, and GPT-3.5 to assess the robustness of the proposed method across varying model architectures and capacities. LLaMA-3 70B is a state-of-the-art open-weight model developed by Meta, offering strong instruction-following capabilities. GPT-4.1-nano is an efficiency-oriented variant of the GPT-4 family, optimized for low-latency deployment with reduced computational cost. GPT-3.5, a widely adopted predecessor to GPT-4, provides a strong performance baseline with balanced accuracy and efficiency. 
As shown in Table~\ref{tab:ablation llm}, GPT-3.5 outperforms the other models, achieving the highest scores in terms of F1, BLEU, and ChrF2 across both translation directions. However, the performance gap relative to the other models is modest, with differences limited to a few percentage points. Moreover, the results are consistent and robust, as evidenced by the low standard deviation observed across all configurations.

\section{Discussion}
\label{sect:discussion}

This study evaluates AulSign for SLT, particularly in low-resource scenarios, using SignWriting as an intermediate representation. AulSign achieves state-of-the-art results for both ASL and LIS in spoken-to-sign and sign-to-spoken tasks, consistently outperforming Jiang et al.~\cite{jiang2022machine} across different data configurations. Notably, AulSign excels in data-scarce environments, demonstrating strong generalization and maintaining performance even with reduced training data.

AulSign is shown to effectively produce linguistically accurate and spatially coherent sign language translations, even with limited data. Its strong performance in sign-to-spoken tasks
% , reflected in high BLEU and chrF2 scores,
is due to its use of context augmentation and domain-specific vocabulary mapping. AulSign adapts well to varying data sizes, maintaining consistent results, unlike the Jiang et al. model, which struggles with smaller datasets. While AulSign is slightly behind the baseline in some data-rich scenarios, its adaptability is superior, especially with smaller or higher-quality datasets.

Across tasks, AulSign demonstrates balanced performance. 
%For ASL, the model achieves average BLEU scores of 22.71 in spoken-to-sign and 24.90 in sign-to-spoken, with chrF2 scores slightly higher in the latter (55.47 vs. 40.07), as shown in Tables 1 and 3.\reminder{è necessario riportare di nuovo i valori qua? Non li abbiamo già discussi alla sezione precedente? LB: e per rafforzare il discorso ma si, li possiamo togliere volendo} 
% \reminder{MM: a quale tabella si fa riferimento? LB: tab 1 e 3}
%This disparity is likely due to the structured prompt in sign-to-spoken, which provides explicit linguistic context for inference.
% \reminder{GT: è vero che probabilmente è dovuto al prompt, perchè facciamo noi sappiamo come le glosse sono combinate, ossia i + per la simultaneità, le parentesi [] per la personificazione etc..., ma non potrebbe essere un sospetto per il reviewer fare intendergli che noi ne conosciamo la loro logica e composizione ? MM: io non ho capito perchè dipende dal prompt. Mi sembrerebbe in realtà dipendere che l'LLM riesce ad esprimersi meglio in inglese piuttosto che in lingua dei segni. Come dire che un madre lingua traduce meglio dalla lingua straniera alla propria piuttosto che viceversa.LB: ok, possiamo eliminarlo o modificarlo.. ditemi voi.}
In LIS translation, AulSign consistently outperforms Jiang et al., benefiting from the highest quality level of the Italian LaCAM CNR-ISTC dataset.
AulSign’s approach leverages a predefined, domain-specific vocabulary, which plays a central role in improving token alignment and translation accuracy. This strategy ensures precise mappings between signs and their representations, particularly in structured intermediate formats like SignWriting. However, signs not present in the vocabulary are treated as unknown, which may affect the model's ability to handle dynamic linguistic contexts or rare signs. Expanding the vocabulary to cover a broader range of linguistic variations and incorporating mechanisms for inferring or adapting to unseen signs would further enhance the model’s capacity to address diverse translation challenges. This is particularly relevant in high-resource settings, where the model's performance is highly sensitive to vocabulary size and coverage, leading to comparable but slightly lower performances in the full-data spoken-to-sign ASL configuration (scenario I).
Finally, AulSign’s modular architecture ensures transparency at each stage of the translation pipeline. The retrieval module, for instance, suitably select in-context samples, while the prompt generation phase explicitly aligns input sentences with grammatical rules. This design allows for targeted error analysis, such as identifying misalignments between FSW symbols and canonical descriptions, which is infeasible in end-to-end SLT models. By prioritizing explainability, AulSign bridges the gap between high performance and user trust, offering a reliable tool for researchers and end-users.

\vspace{-0.05in}

\section{Limitations}
\label{sect:limit}
While AulSign demonstrates promising results, certain limitations affect its generalizability and broader applicability. The model relies on a vocabulary of canonical descriptions to map signs to natural language, ensuring consistency and control over translation quality. However, this dependency introduces a degree of rigidity, as it relies on a specialized vocabulary in which a one-to-one correspondence between signs and their descriptions must be explicitly defined. Furthermore, although AulSign is designed to perform effectively in low-resource settings, its translation quality remains contingent on the availability and consistency of training data. Incomplete or inconsistently annotated datasets may adversely impact accuracy, particularly in capturing the spatial positioning of signs. At present, this study employs SignWriting as the primary intermediate representation due to its structured and expressive nature. Nonetheless, the approach is inherently extensible to alternative notational systems, such as HamNoSys, pose representations, such as SMPL-X, or specialized lexical glossaries, thereby offering potential avenues for enhanced adaptability and integration within automatic sign language translation frameworks. Finally, the current reliance of the Sign Mapper on rigid vocabulary matching could be mitigated through grammar‑constrained decoding methods~\cite{tuccio2025grammar}, which would allow the generation process to directly enforce structural and syntactic constraints, thereby reducing dependence on explicit one‑to‑one mappings.

% \reminder{lb: ricordiamo che dobbiamo stare nelle 7pag+1! (Direi di fare un check eliminando i reminder alla fine)}

\vspace{-0.05in}

\section{Conclusion}
\label{sect:conclusion}
AulSign is a novel framework for sign language translation that leverages retrieval-augmented generative models for enabling translation to previously unseen languages. It achieves state-of-the-art performance in spoken-to-sign and sign-to-spoken translation for both English ASL and Italian LIS, excelling in low-resource scenarios. By using structured vocabularies and a modular, explainable architecture, AulSign enhances translation accuracy, interpretability, and error analysis. %While it relies on predefined vocabularies, expanding lexical coverage and addressing unseen signs remain areas for improvement. 
Future work aims to integrate multimodal data, increase adaptability, and extend support to additional sign languages.

\vspace{-0.05in}

\section{Acknowledgment}
We acknowledge financial support from the Next Generation EU Program with the Future Artificial Intelligence Research (FAIR) project, code PE00000013, CUP 53C22003630006, and by the PNRR project Learning for All (L4ALL) funded by the Italian MIMIT (number: F/310072/01-05/X56). We thank the members of the LaCAM Laboratory, especially Chiara Bonsignori, Alessio Di Renzo, Gabriele Gianfreda, Luca Lamano, Tommaso Lucioli, Barbara Pennacchi, and the LaCAM supervisor, Olga Capirci.

% Italian Research Center on High Performance Computing, Big Data and Quantum Computing (ICSC). We acknowledge financial support from the Learning for All (L4ALL) project, number F/310072/01-05/X56.

\bibliography{custom}

\end{document}